\documentclass[letterpaper,journal]{IEEEtran}
\usepackage{amsmath,amsfonts}
\usepackage{algorithm}
\usepackage{algpseudocode}
\algrenewcommand\algorithmicrequire{\textbf{Input:}}
\algrenewcommand\algorithmicensure{\textbf{Output:}}
\usepackage{array}
\usepackage[caption=false,font=normalsize,labelfont=sf,textfont=sf]{subfig}
\usepackage{textcomp}
\usepackage{stfloats}
\usepackage{breqn}
\usepackage{url}
\usepackage{verbatim}
\usepackage{graphicx}
\usepackage{cite}
\usepackage{blindtext}
\usepackage{mathtools}
\usepackage{optidef}
\usepackage[table]{xcolor}

\def\mclimits_#1{\limits_{\mathclap{#1}}}
\begin{document}

\title{Graph Neural Assisted Actor-Critic for Latency-Efficient Edge Vision System }
\author{\IEEEauthorblockN{Alam Noor\IEEEauthorrefmark{1},
Luis Almeida\IEEEauthorrefmark{2},
Kai Li\IEEEauthorrefmark{3}, 
Jiyan Wu\IEEEauthorrefmark{4},
Miguel Gutiérrez Gaitán\IEEEauthorrefmark{5} and
Eduardo Tovar\IEEEauthorrefmark{1} }

\IEEEauthorblockA{\IEEEauthorrefmark{1}CISTER Research Center, Porto, Portugal.}
\IEEEauthorblockA{\IEEEauthorrefmark{2}Instituto de Telecomunicações, Faculdade de Engenharia, Universidade do Porto, Portugal.}
\IEEEauthorblockA{\IEEEauthorrefmark{3}Department of Information Technology, Kennesaw State University, USA.}
\IEEEauthorblockA{\IEEEauthorrefmark{4}OmniVision Technologies Inc.}
\IEEEauthorblockA{\IEEEauthorrefmark{5}Department of Electrical Engineering, Pontificia Universidad Católica de Chile, Santiago 7820436, Chile.}

}



\maketitle

\begin{abstract}

UAV on-board vision systems are widely used for different activities, including monitoring in no-fly zones. In this case, the vision-equipped UAV streams a video to a ground server where an operator assists its activities. The latency of video transmission has a profound impact on the effectiveness of the operator assistance. However, most techniques available for video transmission still incur significant latency costs. 
In this paper, we propose a graph convolutional neural network-assisted (GCN-assisted A2C) deep reinforcement learning (DRL) system model to find the optimal pixel-correlated area of a suspicious object. We combine the Lagrangian dual form with gradient descent to prevent lack of convergence and over- and under-penalization constraint violation during latency optimization. The proposed system model sends a sub-group pixel-correlated area of the frame from the UAV to the server rather than the transmission of the whole video frame. The proposed framework utilizes the GCN model to explore hidden representations of feature-correlated groups of pixels. Moreover, the GCN supervises the A2C model, which selects a subgroup to enhance transmission latency, thus supervising the training of UAV actions in A2C. Experimental results show that GCN-assisted A2C reduces video frame transmission latency together with false detection rate in UAV vision systems over other DRL and state-of-the-art models.

\end{abstract}


\section{Introduction}
Unmanned Aerial Vehicles (UAVs) are used for a wide spectrum of mobile vision applications, such as augmented reality \cite{10.1145/3281505.3283397}, trajectory planning \cite{10200966}, or visual surveillance systems (e.g., to detect intruders in no-fly zones) \cite{9750351, 10.1145/3561971, 9889286}. Many of these applications rely on real-time  computer vision and require UAVs to detect both large and small objects in dynamic environments with high accuracy. For this purpose, an edge computing approach is common, having the UAVs stream video to a server in a ground station for assistance, getting important data from the server after preprocessing to identify objects with high accuracy \cite{9488704}\cite{10274950}. However, the transmission of video frames with high resolution imposes long delays that hinder real-time operation. Similarly, doing local video processing onboard the UAV is highly constrained due to limited computing resources, imposing long delays and/or low video resolution~\cite{9780408}. 


A promising solution to combine low latency and high accuracy detection following the edge computing paradigm is to offload to an edge server  equipped with GPUs running deep learning models the processing of just specific areas of the video frames representing regions-of-interest (RoI) \cite{9488843, 9912279, 9796984}. This reduces the amount of information to be transferred, leading to lower transmission time, thus reducing the latency of the Edge support. 
However, user-side parameters like image resolution and encoding rate and server-side parameters like neural network architecture have a significant impact on video analysis if we do not consider how the pixels of the RoI correlate with the pixels of the video frame background environment \cite{9912279}. 

\begin{figure}[htb]
    \begin{center}
    \hspace*{0em}
        \includegraphics[scale=0.8]{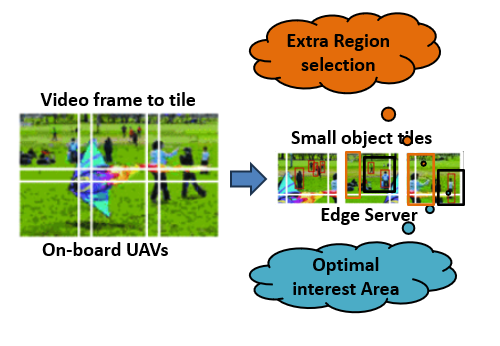}
        \caption{EdgeDuet non-optimal region selection for the transmission of small objects to an Edge server.}
        \label{EdgeDuet}
    \end{center}
\end{figure}

Wang et al. \cite{9488843} studied a split-based object detection model (EdgeDuet) between UAVs and the Edge. The detection of small objects is offloaded to the server using an RoI of the frame and content-prioritized tiles. The RoI of the frame transmitted with pixel blocks potentially contains small objects of high resolution. The rest of the frame has large objects that can be detected with low resolution onboard the UAV. However, EdgeDuet is based on tile selection, with tiles having an area that is unrelated to the size of the small objects, thus potentially causing the transmission of more information than strictly needed, incurring unnecessarily longer transmission latency. Moreover, as shown in Fig.\ref{EdgeDuet}, the use of tiles may also hinder the accuracy of server detection. 

Yang et al. \cite{9796984} introduced the Flexpatch model that allows for the detection of small objects with a small delay while also addressing occlusion, changes in object appearance, and the presence of additional objects. However, the Flexpatch model is based on optical flow with static cameras, thus unsuited for moving cameras onboard UAVs.

\begin{figure}[ht]
    \begin{center}
    \hspace*{0em}
        \includegraphics[scale=0.45]{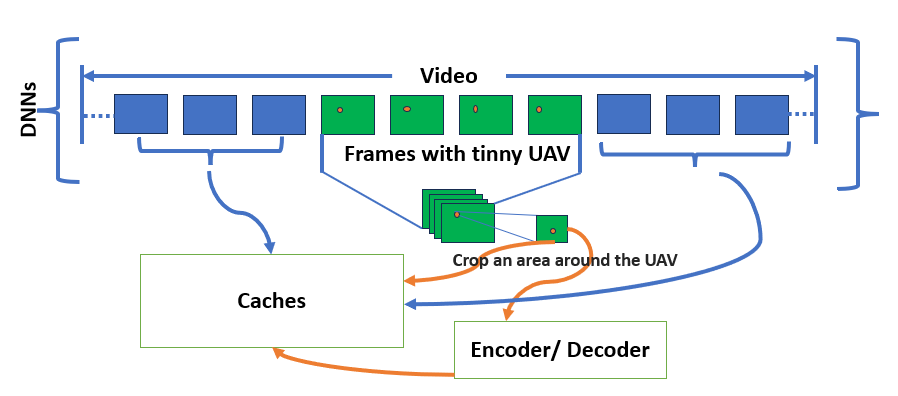}
        \caption{Onboard UAV processing video to detect suspicious UAVs.}
        \label{On_board_processing}
    \end{center}
\end{figure}

In this work we aim at managing the sizes of the sub-area video frames to reduce the amount of information to be transferred and the associated transmission delay (Fig.~\ref{On_board_processing}). The key of the system model is a Graph Convolutional Network (GCN) that selects from a video frame the RoI of small objects and the group of neighboring pixel regions that are correlated with the centroid of the RoI. Then, it is important to select the group of pixels-correlated regions from the output of the GCN module that optimizes the latency subject to satisfying the accuracy of object identification.

For this purpose, we propose a novel GCN-assisted Advantage Actor Critic Deep Reinforcement Learning model (GCN-assisted A2C) that is fast, accurate, and flexible to accommodate the movement of the onboard UAV camera. 
Our objective is to efficiently optimize and control the transmission delay in order to accurately identify intruding UAVs in a no-fly zone. We use the YOLO-Nano model to define the bounding box for detected UAVs. The GCN uses the bounding box center to find the group of pixels-correlated regions with a strong relationship with the RoI centroid. The A2C module of the proposed GCN-assisted A2C model selects the actual group of pixels-correlated regions to be sent to the server. Finally, the server finds the object and sends back the information of accuracy and latency for the GCN-assisted A2C model to optimize the latency efficiently, reducing the transmission delay by selecting a minimum group of pixels-correlated regions that allows meeting a given target accuracy.

In summary, the main contribution of this paper is a novel GCN-assisted A2C model that encompasses the following two parts:
\begin{itemize}
    \item 
    A GCN that uses the bounding boxes of RoI produced by an onboard YOLO-Nano algorithm and creates clusters of feature-correlated groups of pixels and regions of hidden pixel relationships. The GCN model is trained with a SLIC segmentation of frames as ground truth for the groups of pixel regions.
    
    \item 
    An A2C that uses the GCN as a state based on a group of pixels correlation values of future predictions to process the output, take action, and transmit it to the server. The A2C reward is based on the Lagrangian dual form for checking the sensitivity of the objective function. The aim is to reduce the risk of temporal difference learning error estimation for both the critic and the actor loss to update the policy. We use dual gradient descent to update the $\lambda$ parameter and avoid lack of latency convergence and over- and under-penalization constraint violation during latency optimization.
\end{itemize}

We trained the proposed GCN-assisted A2C model on a YouTube dataset to assess its real-world performance and compared it with three benchmark models: FlexPatch, EdgeDuet, and DRL. The proposed model achieved a transmission latency of 45~ms, approximately half of what was achieved with the competing approaches. It also achieved an average precision of 70.72\%, with a mean Intersection over Union (IoU) of 60.30\%, indicating more accurate RoI selection and higher overall detection performance when compared against the benchmark methods.

Beyond the proposed GCN-assisted A2C, this work uses the following technologies in the end-to-end video pipeline: 
\begin{itemize}
    \item \textbf{UAV side:} An onboard camera combined with YOLO-Nano inference for real-time object detection. When the detection confidence falls below a predefined threshold, region-of-interest (RoI) image crops are extracted and transmitted to the edge server using an intra-coded image encoding scheme (H.264 Intra-only), enabling efficient and robust transmission of independent RoI frames.
    
    \item \textbf{Edge server side:} ESRGAN to regenerate missing contextual information, FFDNet for image denoising, Zero-DCE for low-light enhancement, followed by a large-scale YOLO model for accurate object detection and refinement.
\end{itemize}

The rest of the paper is organized as follows. Section II discusses related work, Section III presents the mathematical system modeling, Section IV introduces the GCN-assisted A2C system model, and Section V validates the proposed model with an experimental study and discusses the results. Finally, Section VI concludes the paper.

\section{Related Work}
Monitoring no-fly zones for suspicious flying objects, particularly UAVs, can be achieved by resorting to a legal UAV that captures and streams live video from the air. However, this approach requires a significant volume of live video transmission from the UAV to an Edge server, consuming computing resources and incurring high transmission latency \cite{10.1145/3641282}\cite{9941155}. To address these challenges, the previous system models might be categorized in different classes based on their primary approach to minimizing the transmission latency and improving the performance of real-time detection of UAVs.

Several researchers introduced low-latency video streaming system models to reduce video transmission latency and improve real-time detection. Dong et al. \cite{9780408} introduced an ultra-low latency frame delivery model using the UDP-based QUIC protocol and the widely compatible TS format, achieving latency below 200ms. Wang et al. \cite{8901872} studied feature-based video transmission, using Lagrangian dual decomposition to optimize the transmission latency and semantic image segmentation for video feature selection. Qu et al. \cite{10.1007/978-3-031-61060-8_24} studied UAV-swarm-aided modeling based on a disaster response platform for the video transmission from UAVs to the ground as air-to-ground coordination, while Khan et al. \cite{10381707} presented a public safety communication network model that uses observation UAVs to receive video streams from ground users in an affected area and transmit them to a nearby supporting ground station.

Some works focused on combining edge and cloud computation to optimize latency. Anurag et al. \cite{10.1145/3576842.3582385} studied edge and cloud collaborative work to use heavy deep learning models at the cloud and small models at the edge to fuse their predictions to optimize latency.  Liu et al. \cite{LIU2024103443} designed a multi-approach model that reduces latency by extending splitting-merging streaming  and multi-link retransmission to reduce the packet loss. Yaqoob et al. \cite{10536624} presented a DRL-based soft actor-critic model to improve the quality of experience of video transmission latency from a UAV to a ground server. Meanwhile, Song et al. \cite{10273375} introduced an AI-driven, multipath transmission of a UAV live streaming system to optimize bandwidth aggregation and transmission reliability, while Wu et al. introduced a multiobjective control strategy learning for a UAV using a soft actor-critic DRL algorithm to manage the unstable transmission quality problem of videos in a dynamic topology network~\cite{10361535}. Moreover, UAV--MEC studies apply deep reinforcement learning directly to video and task offloading. Zhao et al.~\cite{10185964} jointly optimize resolution, offloading decision, and resource allocation for secure video offloading in multi-UAV MEC networks, while Lee et al.~\cite{10186448} and Lakew et al.~\cite{9743399} adopt multi-agent reinforcement learning to control the offloading ratio and resource allocation in UAV and aerial-IoT edge networks. Closer to the detection setting, Li et al.~\cite{10239290} offload UAV object detection and segmentation to an edge server for fast, low-latency inference.

Other approaches focus on detecting objects efficiently by RoI and object-centric models to transmit video to relevant areas. Zhang et al. \cite{9912279} worked on an instance segmentation technique executed on the Edge (EdgeIS) to segment an RoI and transfer it to a mobile device with low latency. The EdgeIS model detects objects and tracks their motions with mask transferring. Moreover, EdgeIs models utilize the contour-instructed edge inference acceleration scheme to reduce latency. Wang et al.~\cite{9488843} studied EdgeDuet, a split-based object detection framework between UAVs and the Edge, where regions of interest (RoIs) containing small objects are transmitted to an Edge server using content-prioritized tiles, while the remaining regions are processed onboard at lower resolution. Although effective, EdgeDuet relies on tile-based RoIs whose areas are not aligned with the actual size of small objects, potentially leading to unnecessary data transmission, increased latency, and reduced detection accuracy. Yang et al.~\cite{9796984} introduced Flexpatch, an RoI-based approach that enables low-latency detection of small objects by tracking and transmitting adaptive patches corresponding to regions of interest, while being robust to occlusion and appearance changes; however, Flexpatch relies on optical flow with static cameras, making it unsuitable for moving cameras onboard UAVs.

Among the works referred to above, only a few aim at reducing the video transmission latency for the specific purpose of flying objects real-time detection \cite{9780408,9488843,9912279,9796984}. Some of the works \cite{9488843,9912279,9796984}, though, still stress the use of the network channel bandwidth transmitting areas of the video frames of no interest for flying object detection. This fact limits their capacity to reduce video latency and packet losses since there will be more packets transmitted than strictly needed. 

In our work we pursue an approach based on identifying regions of interest in the video frames and transmit just these RoIs, leaving to the Edge server the task of accurately identifying small objects in those RoIs, even with poor video quality or environmental effects. The works that best relate to ours are EdgeDuet \cite{9488843} and Flexpatch \cite{9796984}, which we will use for comparison. Our work is, to the best of our knowledge, the only one that trades off Average Precision (AP) with latency, using an A2C DRL approach to get the lowest latency (by means of selecting the smallest RoIs) that allows meeting a given AP target.









\section{Mathematical System Modeling}
\label{sec:model}
We consider a sequence of encoded video frames $\{F_t\}$ of spatial size $W\times H$ pixels captured by a UAV. A light onboard detector (e.g., YOLO-Nano) produces at times $t$ a bounding box $b_t = (x_c,y_c,\zeta_W,\zeta_H)$ with center $(x_c,y_c)$ and size $(\zeta_W,\zeta_H)$. $\zeta_W$ and $\zeta_H$ are the width and height of the detected bounding box. The transmitter chooses an action $E_t$ that parameterizes a region of interest $ f_t(E_t)$ centered at $(x_c,y_c)$ with scale factors $\alpha_t$ and $\beta_t$ (Eq.~\ref{eq_delta_ft}). 

\begin{multline} 
\label{eq_delta_ft}
f_t(E_t) = \\
\left[x_c-\tfrac{\alpha_t \zeta_W}{2}, x_c+\tfrac{\alpha_t \zeta_W}{2}\right]
\times \left[y_c-\tfrac{\beta_t \zeta_H}{2}, y_c+\tfrac{\beta_t \zeta_H}{2}\right]
\end{multline}

Let $|f_t|=\alpha_t \beta_t \zeta_W \zeta_H$ be the RoI area in pixels. For symmetric expansion, we consider $\alpha_t = \beta_t= 1+2E_t$, where $E_t$ is the scaling factor that determines the cropping region of the bounding box size, with $E_{\text{min}} \leq  E_{t} \leq  E_{\text{max}}$. 

In practice, each frame $F_t$ will contain several sub-areas of interest (i.e., the RoIs). Let $\mathcal{F}_t = \{ f_{t,1}, \ldots, f_{t,n}\}$ be the set of such sub-areas in which $f_{t,n}$ denotes the $n$-th sub-area of frame $F_t$ and $N$ the set size. 
Once the $N$ RoIs are defined, they are transmitted through the communication channel. Of particular importance to the transmission time is the RoIs area, which is given by $|f_{t,n}| = (1+2 E_{t,n})^2 \, \zeta_W \zeta_H$, because it determines the amount of information to be transmitted. Without loss of generality, we consider that each RoI is transmitted inside one communication packet of size $S_{t,n}$ as given by Eq.~\ref{eq_size_p}, where $\rho$ (bits/pixel) captures the codec rate at the chosen quality and $\Psi_{\text{frame}}$ aggregates headers and per-packet overhead.\footnote{If tiling is used, 
$\Psi_{\text{frame}}$ can include tile headers; $\rho$ can vary mildly with texture/motion.} 

\begin{equation}
\label{eq_size_p}
S_{t,n} \;=\; 
\underbrace{\rho \, |f_{t,n}|}_{\text{content payload}} 
\;+\; \underbrace{\Psi_{\text{frame}}}_{\text{headers/overhead}}
\end{equation}

In real-time the communication between the UAV and the Edge server is affected by packet loss, which in turn triggers retransmissions, impacting the RoIs transmission times. This loss-induced extension of the transmission times must be considered in the target delay constraint \cite{8359128}. 

Therefore, we define the expected transmission time per packet $T_{t,n}$ as given by Eq. \ref{eq_expected_tx} \cite{10065636}\cite{7756386}, where 
$R$ (bits/s) is the wireless link transmission rate, $\min RTT$ is the round-trip time, and $q_{t,n}$ is the per-packet success probability (after PHY/MAC). We also consider a stop-and-wait ARQ retransmissions protocol with a NACK packet of size $B$ bits. 

\begin{dmath}
\label{eq_expected_tx}
T_{t,n} 
= \underbrace{\left(\frac{S_{t,n}}{R} 
+ \frac{\min RTT}{2}\right)}_{\text{first attempt}} 
+ \\
+ \underbrace{\frac{(1 - q_{t,n})}{q_{t,n}}}_{\text{expected retries}}
\cdot
\underbrace{\left(\frac{B}{R} 
+ \min RTT 
+ \frac{S_{t,n}}{R} \right)}_{\text{each retry}},
\end{dmath}

\subsection{Accuracy Constraints Strategy}
We now construct the constraints for our framework regarding detection accuracy and sub-area frame size, emphasizing the importance of carefully selecting appropriate frame cropping area scales for UAV detection accuracy and how it greatly influences performance. The necessity for careful sub-area frame scale selection arises from two primary factors: first, real-world UAVs come in a wide range of sizes, often unfamiliar to the visual system; and second, the distances between objects and the camera may vary, potentially lacking prior information. We aim to achieve sufficiently high accuracy $(\mathcal{A}_{t,n})$ on the server side when receiving sub-area frame $f_{t,n}$. 

We consider the server side accuracy $\mathcal{A}_{t,n}$ to be a function of RoI size as shown in (\ref{Accuracy}). 
Note that $\mathcal{P}\{\cdot\}$ is a probability of each transmitted sub-area, namely the likelihood that the server correctly detects an intruding UAV given the received RoI, i.e., $\mathcal{P}\{\text{correct detection} \mid f_{t,n}\}$.

We use (\ref{Accuracy}) to estimate on the inspecting UAV side the accuracy on the server side and decide on the need to scale the RoI before transmission so that a desired server side accuracy threshold $A_{\text{threshold}}$ is met. This procedure is shown in (\ref{condition_accuracy}) in which we scale up the sub-area frame choosing an optimal scaling factor $E_{t,n}^{(k+1)}$ iteratively  until the threshold is overcome, starting from $E_{t,n}^{(0)}=0$ for $k=0$.



\begin{equation}\label{Accuracy}
\mathcal{A}_{t,n}^{(k)} \;=\; 
\mathcal{P}\!\big(| f_{t,n}^{(k)}|\big)
\;=\;
\mathcal{P}\!\Big((1+2 E_{t,n}^{(k)})^2\,\zeta_W\,\zeta_H\Big).
\end{equation}

\begin{equation}\label{condition_accuracy}
\mathcal{A}_{t,n}^{(k+1)} =
\begin{cases}
\mathcal{P}\!\big(|f_{t,n}^{(k+1)}|\big), & \text{if } \mathcal{A}_{t,n}^{(k)} \le A_{\text{threshold}}, \\[6pt]
\mathcal{A}_{t,n}^{(k)}, & \text{otherwise}.
\end{cases}
\end{equation}


The scaling factor is updated as $E_{t,n}^{(k+1)} = V_{E} \cdot E_{t,n}^{(k)}$ where $V_{ E}$ is computed proportionally to the correlation of the group of pixels of the sub-area with respect to its center. 

Moreover, $\mathcal{P}\{\cdot\}$ must be optimal to obtain maximum accuracy and low latency. A retrieval process based on $E_{t,n}$, treated as a ranking or sorting method for different sub-area frames, does not work because 
it is non-differentiable. In addition, linear optimization cannot be applied, since constant changes in the input distance $|f_{t,n}|$ cause the $\mathcal{A}_{t,n}$ values to be discontinuous \cite{8578167}. In order to obtain an optimal $\mathcal{P}\{\cdot\}$, groups of pixel correlations between regions must be considered, since correlations between center and neighboring pixels are essential. This approach also reduces latency by enabling a one-time prediction of a sub-area of the frame rather than relying on a sequence of actions. The GNN primarily focuses on prediction and determining the optimal decision for a single occurrence. Using the GNN algorithm to identify 
the area according to pixel correlations results in a more significant decrease in latency compared to linear optimization. This is because linear optimization is based only on $E_{t,n}$, which does not guarantee higher precision in the first transmission of sub-area frames and is vulnerable to selecting inappropriate regions of the frame with low pixel correlations.

\subsection{Problem Formulation} 
The above analysis shows that the number of transmissions of sub-area frames $f_{t,n}$ in $T_{t,n}$ is critical to maintaining the delay below the desired threshold of a single full-frame transmission delay. According to the \cite{10.1145/3485447.3512276}, the total transmission time of frames can be expressed as the sum of all single sub-area frame delays, which is given in (\ref{time}).

\begin{dmath}\label{time}
\sum_{n=1}^{N} T_{t,n} 
= \sum_{n=1}^{N} \Biggl(
   \frac{S_{t,n}}{R} 
   + \frac{\min RTT}{2} +\\
   + \frac{(1 - q_{t,n})}{q_{t,n}}
     \left(\frac{B}{R} 
     + \min RTT 
     + \frac{S_{t,n}}{R}\right)
   \Biggr)
\end{dmath}

The optimization problem to minimize the sum of the total delay can be stated as follows. Given the estimated packet sizes $S_{t,n}$ for each transmission at send rate $R$ and the one-way propagation/processing delay $\frac{\min RTT}{2}$, the objective is to achieve minimum transmission latency by adapting each cropped sub-area $f_{t,n}$ via a variable scaling factor $E_{t,n}$ per packet. The design also accounts for the number of sub-area transmissions and leverages an optimal $\mathcal{P}(.)$ (pixel-correlation-based precision probability), so that the server-side 
accuracy $\mathcal{A}_{t,n}$ for each received sub-area frame is high and stable. Formally,

\begin{dmath}\label{OP}
\textbf{OP} :\quad 
\arg \min_{\{f_{t,n}\}_{n=1}^N} \ \sum_{n=1}^N T_{t,n}
\end{dmath}

\begin{equation*}
\text{subject to:} \quad 
\mathcal{A}_{t,n} \geq A_{\text{threshold}}, \quad n=1,\ldots,N. 
\end{equation*}

where

\begin{equation*}
\begin{split}
T_{t,n} 
&= \left(\frac{S_{t,n}}{R} 
   + \frac{\min RTT}{2}\right) +\\
   &+ \frac{1 - q_{t,n}}{q_{t,n}}
    \quad \left(\frac{B}{R} + \min RTT 
   + \frac{S_{t,n}}{R}\right), \\[0pt]
S_{t,n} 
&= \rho \, |f_{t,n}| + \Psi_{\text{frame}}, \\[0pt]
\mathcal{A}_{t,n} &= \mathcal{P}\!\big(|f_{t,n}|\big),\\[0pt]
 E^{k+1}_{t,n} 
&= V_{ E} \cdot E^{k}_{t,n},\quad E^{0}_{t,n}=0 
\end{split}
\end{equation*}

\subsection{Complexity}
Problem \textbf{(OP)} induces a constrained Markov decision process (MDP) with continuous action variable $ E_{t,n}$ and a non-convex, data-driven accuracy function ${\mathcal{A}_{t,n}}(\cdot)$. Exact dynamic programming is intractable. Therefore, we adopt a model-free primal–dual actor–critic method that learns policies minimizing the expected delay while satisfying accuracy constraints. In the simplified case where $E_{t,n}$ takes values on a discrete grid of $K$ levels, and both $\mathcal{A}_{t,n}$ and $T_{t,n}$ are monotone in $ E_{t,n}$, each sub-area frame can be optimized independently in $O(NK)$ by scanning the grid. However, when temporal correlations, joint constraints, and stochastic packet success are considered, the problem becomes a high-dimensional constrained MDP (CMDP), making reinforcement learning methods more appropriate.

\section{Proposed GCN-Assisted DRL System Model}
This section presents the model of the GCN-assisted A2C system. The proposed model employs GCN to extract the areas of features that have a strong relationship with the object center, as well as the strong correlation between groups of pixels to facilitate future prediction. While A2C exploits the learning outcomes of GCN to train the action of latency optimization and accuracy constraint satisfaction. 

In the previous section, we formulated the transmission optimization problem (OP) by defining the action as the RoI scaling factor $E_{t,n}$, which directly 
adjusts the size of the cropped sub-area frame $f_{t,n}$. While this abstraction provides a tractable CMDP formulation, it does not exploit correlations between pixels inside the bounding box, nor does it account for complex feature distributions that arise under different UAV sizes, lighting conditions, and motion. To address these limitations, we extend the baseline model with a GCN-assisted A2C framework. In this setting, the GCN predicts correlated pixel groups $\mathbb{S}_i$ that summarize feature relationships in the frame, and these groups are mapped into 
bounding boxes $\mathbb{B}_{t,n}$ centered around correlation-weighted centroids. Each $\mathbb{B}_{t,n}$ corresponds to an RoI that can be expressed as $f_{t,n}$ for some effective scaling factor $E_{t,n}$. In other words, the GCN refines the baseline $E_{t,n}$-based action by constraining the A2C policy to choose feature-correlated regions that are both latency-efficient and accuracy-preserving. The following presents the proposed GCN–A2C solution model in detail. 

\subsection{CMDP Formulation and Information Flow}
At each time step $t$, the CMDP state is defined as $s_t=\big(\phi_t^{\mathrm{det}},\, \mathbf{g}_t,\, q_{t},\, \min RTT_t,\, \lambda_t\big)$,
where $\phi_t^{\mathrm{det}}$ are detector features (bounding boxes, scores), $\mathbf{g}_t$ is a GCN embedding summarizing pixel-group correlations around $(x_c,y_c)$, and $\lambda_t$ is the CMDP multiplier for latency–accuracy constraints.\footnote{Not all state variables appear directly in the OP formulation, which is expressed only in terms of the optimization variables $f_{t,n}$, $E_{t,n}$, $T_{t,n}$, and 
$\mathcal{A}_{t,n}$. Instead, the CMDP state aggregates the contextual information needed by the RL policy to generate decisions $f_{t,n}$ that solve OP in practice.} 

In the proposed system model, the feature vector $(\mathbb{V})$ for GCN obtained from pixel-group sets $\mathbb{S}=\{\mathbb{S}_i\}_{i=1}^Z$ to form the GCN embedding $\mathbf{g}_t$. That is, $\mathbf{g}_t = \mathbb{V}$ provides a compact encoding of correlations 
between disjoint pixel groups, which the A2C agent uses to guide decision-making. 

The CMDP action corresponds to selecting a bounding box $\mathbb{B}_{t,n}$ from the GCN-predicted correlated pixel group $\mathbb{S}^*_i$. Each bounding box $\mathbb{B}_{t,n}$ 
defines a sub-area RoI in the original OP formulation:
$\mathbb{B}_{t,n} \equiv f_{t,n}$. Thus, the GCN transforms pixel correlations into bounding-box actions that the A2C policy transmits to the server, balancing 
latency $T_{t,n}$ and accuracy $\mathcal{A}_{t,n}$. After transmission, the server returns the observed accuracy (AP or IoU proxy) 
and measured delay, which update the estimated accuracy ${\mathcal{A}}_s$ and success probability ${q}_t$ online.

\subsection{GCN-A2C for Decision Making}
We use an A2C strategy to optimize latency for solving the OP problem formulated from (\ref{eq_size_p}) to (\ref{time}) \cite{9793853}. Figure~\ref{proposedmodel} illustrates a framework with three primary modules. The first is the \textbf{actor module}, which takes an action $\{\mathbb{B}_{t,n}, \mathbb{S}_i\}$. Here, the GCN predicts correlated pixel groups $\mathbb{S}_i$, and the corresponding bounding box $\mathbb{B}_{t,n}$ defines the cropped sub-area. Formally, each $\mathbb{B}_{t,n}$ corresponds to an RoI $f_{t,n}$ with scaling factor $E_{t,n}$, i.e., $\mathbb{B}_{t,n} \equiv f_{t,n}$. Without GCN, the A2C actor 
would need to search exhaustively over pixel groups, leading to high computation. The pixel-group shape adapts to the variance or distribution around the centroid and has width $\zeta_W$ and height $\zeta_H$. The \textbf{critic module} evaluates $\{T_{t,n}, \mathcal{A}_{t,n}\}$, i.e., the transmission delay per sub-area packet and the accuracy (mAP proxy) achieved at the server. The \textbf{policy} is improved by combining A2C training 
with GCN predictions: the GCN records correlations in pixel groups and provides future predictions, which periodically update the actor to select feature-correlated regions that minimize latency while ensuring $\mathcal{A}_{t,n}$ constraints are met. This guarantees that the policy $\pi(a_t|s_t)$ adapts to varying UAV sizes and pixel correlations. The size of each pixel group is determined by $\Theta = \sqrt{\frac{K}{M_i}}$, where $K$ is the total number of correlated pixels near the object center, and $M_i = |\mathbb{S}_i|$ is the cardinality of the subset used by the GCN to guide 
actor decisions. Exhaustively evaluating groups of size $M_i$ requires $\binom{K}{M_i} = \frac{K!}{M_i!(K-M_i)!} = O(K^{M_i})$, making brute-force search infeasible. Instead, the GCN learns to predict future feature-correlated actions 
for A2C, reducing complexity while satisfying $\mathcal{A}_{t,n}$. The proposed system not only selects pixel groups but also captures stronger correlations between the center and neighboring hidden pixels. In the next section, 
we detail the structure of the GCN model.

\subsection{Hidden Pixels Relation Exploration}
\begin{figure*}[ht]
    \begin{center}
    \hspace*{0em}
        \includegraphics[scale=0.4]{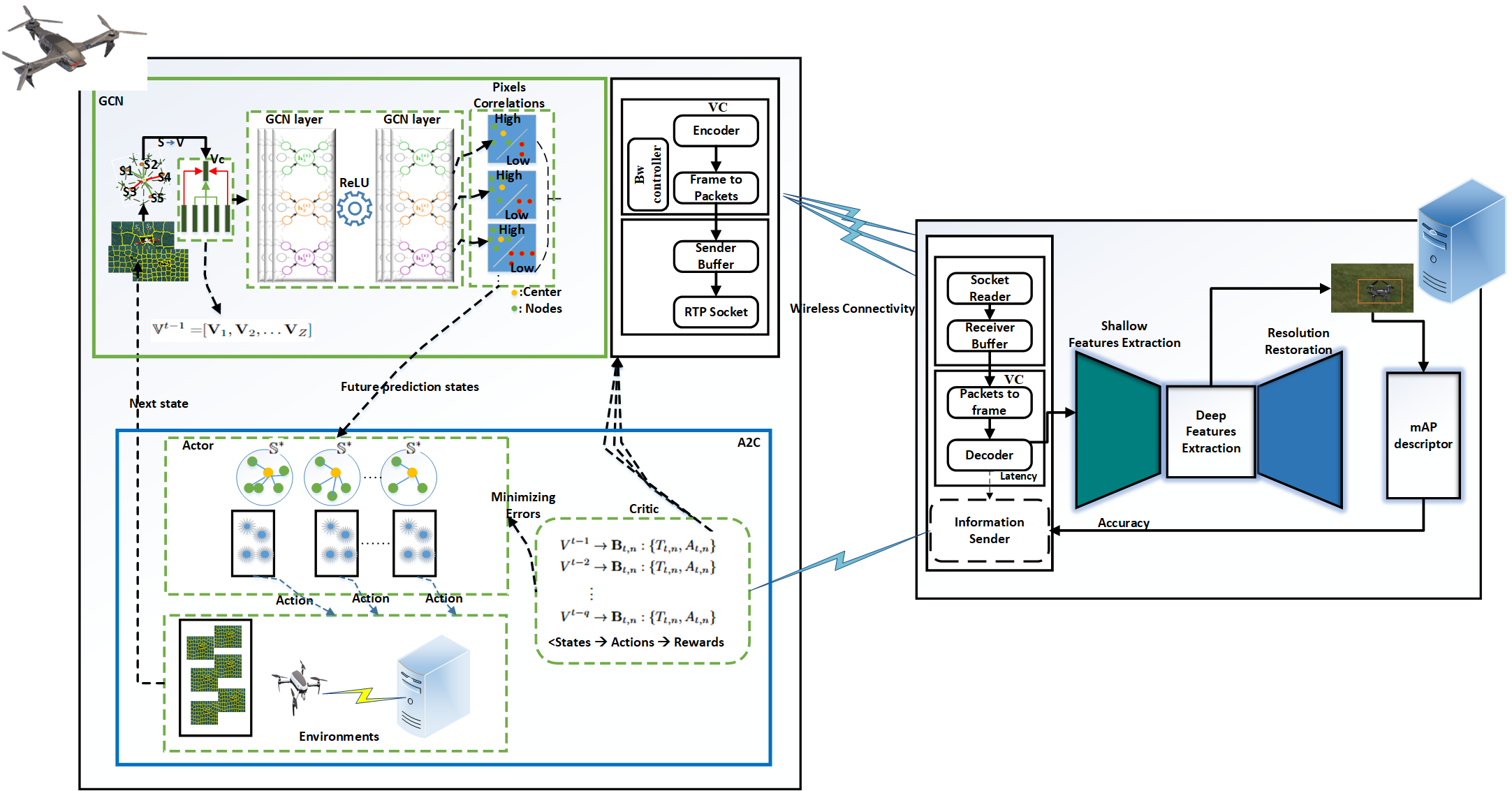}
        \caption{The figure illustrates the proposed framework, indicated by several color-coded parts. The left-side black box encloses the proposed model region, which symbolizes the core framework of the GCN-assisted A2C. The blue box shows the actor who selected the action. While the critic's role in action-value evaluations. The green box is the GCN future state prediction, which regulates actor behavior to optimize latency and minimize errors.}
        \label{proposedmodel}
    \end{center}
\end{figure*}

This section presents a detailed graphic representation along with A2C of the proposed graph neural network structure for higher accuracy and low latency, shown in Fig. \ref{proposedmodel}. To find the high correlations between the pixels of an image for optimal sub-area frame selection, we need to divide the frame into different partitions. The frame partitioning process is a critical and fundamental stage within any low-level vision system to find the region of interest where pixels have higher correlations with each other. Segmentation refers to the procedure of dividing a frame into distinct areas that do not overlap, with the objective of ensuring that each zone shows correlations and that the separation between areas is abrupt. Different methods for pixel correlations have been published in the past. However, it remains a challenge to identify a specific approach that shows consistency throughout numerous lighting conditions or low-visibility conditions. In addition, CNN performs convolution operations in smaller, more uniform areas and obtains features at the pixel level. However, GCN performs convolution operations in large areas with unevenly correlated pixels and gets features at the aggregated pixel level. The GCN method uses the average of all pixels in a set to give strong edge weights to areas with similar correlations and weak or no edge weights to areas with no correlations \cite{10200966}\cite{9793853}.

\subsubsection{State}
Let $\mathbb{S} = \{\mathbb{S}_i\}_{i=1}^Z$ represent the set of aggregate pixel groups forming the state, where $\mathbb{S}_i = \{\mathbf{Y}_i^j\}_{j=1}^{M_i}, \qquad M_i = |\mathbb{S}_i|$. Here, $\mathbb{S}_i$ denotes the $i$-th group of correlated pixels and $M_i$ is its cardinality. The pixel groups are disjoint, i.e., $\mathbb{S}_i \cap \mathbb{S}_j = \emptyset$ for all $i \neq j$. Moreover, the 
total number of pixels is preserved: $ \zeta_H \times \zeta_W = \sum_{i=1}^{Z} M_i$. Each group is summarized by its mean feature vector, yielding the state 
representation, which is given in (\ref{feature_vector}): 

\begin{dmath}\label{feature_vector}
    \mathbb{V} 
    = {\left[ {{\mathbf{V}}_{1}}, {{\mathbf{V}}_{2}}, \ldots, {{\mathbf{V}}_{Z}} \right]^{T}} \\
    = {\left[ {\frac{1}{M_1} \sum_{j=1}^{M_1} \mathbf{Y}_1^j, \dots, 
    \frac{1}{M_Z} \sum_{j=1}^{M_Z} \mathbf{Y}_Z^j } \right]^{T}}, 
\end{dmath}
where $\mathbf{Y}_i^j$ is the $j$-th pixel feature in group $\mathbb{S}_i$. Thus, $\mathbb{V}$ is a column vector of dimension $Z$, with each entry being the average feature of a pixel group. This representation captures the correlation structure of the aggregated pixel regions, which serve as the 
input nodes for the GCN. After processing the initial state, the GCN predicts future correlation patterns, which guide action selection by cropping the corresponding bounding box region. A focal classification loss $\mathfrak{L}_{cls}$ is used to train the GCN for UAV pixel correlation prediction, which is given in (\ref{classification_loss}):
\begin{equation}\label{classification_loss}
\mathfrak{L}_{cls} = \lambda_{1}\mathfrak{L}_{tcrp} + \lambda_{2}\mathfrak{L}_{tcrn},
\end{equation}
where $\mathfrak{L}_{tcrp}$ measures correlation consistency among pixels inside the UAV bounding box, while $\mathfrak{L}_{tcrn}$ penalizes spurious correlations in background regions outside the UAV. The loss $\mathfrak{L}_{tcrn}$ suppresses background noise near the UAV centroid, allowing the GCN to emphasize UAV-related pixels. Here, $\lambda_1$ and $\lambda_2$ are tunable weighting coefficients used to balance the influence of $\mathfrak{L}_{tcrp}$ and $\mathfrak{L}_{tcrn}$ during training.

\subsection{A2C for $T_{ f}$ Optimization with $\mathcal{A}_{t,n}$ Satisfaction}
\subsubsection{Action}
In the previous section, we proposed a GCN method to capture correlations between pixels by partitioning images into groups of pixels. Using this foundation, we now leverage the correlation data to define the \textbf{action} in the A2C framework, which specifies the region to be transmitted. Let the action be a bounding box $\mathbb{B}_{t,n}$ associated with a correlated pixel group $\mathbb{S}^*_{i}$ predicted by the GCN. The goal is for the GCN to identify highly correlated pixel groups that can be mapped into compact rectangular or square regions. To achieve this, we define a correlation-weighted centroid $(x_c,y_c)$ based on the coordinates $(x_{\Phi},y_{\Phi})$ of pixels ${\Phi} \in \mathbb{S}^*_{i}$ with weights 
$\mathfrak{C}_{\Phi}$:

\begin{equation}\label{x_c}
x_c = \frac{\sum_{{\Phi} \in \mathbb{S}^*_i} \mathfrak{C}_{\Phi} \cdot x_{\Phi}}{\sum_{{\Phi} \in \mathbb{S}^*_i} \mathfrak{C}_{\Phi}}, 
\quad
y_c = \frac{\sum_{{\Phi} \in \mathbb{S}^*_i} \mathfrak{C}_{\Phi} \cdot y_{\Phi}}{\sum_{{\Phi} \in \mathbb{S}^*_i} \mathfrak{C}_{\Phi}}.
\end{equation}

In (\ref{x_c}) $\mathfrak{C}_{\Phi}$ denotes the correlation weight of pixel $\Phi$, highlighting its contribution to the centroid. Using the centroid, the bounding box action $\mathbb{B}_{t,n}$ is defined as:

\begin{equation}\label{B_i}
\mathbb{B}_{t,n}: 
\left[x_c - \tfrac{{\zeta}^*_W}{2},\; y_c - \tfrac{{\zeta}^*_H}{2},\; 
      x_c + \tfrac{{\zeta}^*_W}{2},\; y_c + \tfrac{{\zeta}^*_H}{2}\right],
\end{equation}

In (\ref{B_i}) ${\zeta}^*_W$ and ${\zeta}^*_H$ denote the width and height of the correlated region. These dimensions are adaptive, depending on the variability of the pixel-group distribution around the centroid. This flexibility improves 
the expressiveness of the bounding box and ensures that correlated regions are captured accurately.

\subsubsection{Reward}
The A2C training reward $(r_t^{\text{A2C}})$ determined by the totals of sub-area frames transferred to the server within the action on the group of pixel regions as $\left \{{\mathbb{S}^\psi_i \mid \mathbb{S}^*_i, \mathbb{B}_{t,n}}\right \}$, as $\mathbb{S}^*_i$ represents the correlated group of GCN-trained pixels. The action $\mathbb{B}_{t,n}$ is taken by the UAV, and the GCN-predicted group of pixels transits from 
$\mathbb{S}^*_i$ to $\mathbb{S}_i$. However, the reward can only be optimized when $\mathcal{A}_{t,n} \geq A_{\text{threshold}}$ accuracy is considered for positive detection of the object on the server. It is essential to achieve optimal transmission latency and avoid penalization for constraint violation. It might be possible to put a condition for $\left \{{\mathbb{S}^\psi_i \mid \mathbb{S}^*_i, \mathbb{B}_{t,n}}\right \}$ preserving the feature-correlated region that can be achieved by Lagrangian dual form, as given in (\ref{Lagrangian}):

\begin{dmath}\label{Lagrangian}
    \mathcal{L}(\mathbb{S}^\psi_i, \mathbb{S}^*_i, \mathbb{B}_{t,n}, \lambda) 
    = \min _{\forall i_{\mathbb{S}^*_i} \in [1,Z]} T_{t,n} 
    + \lambda(A_{\text{threshold}} - \mathcal{A}_{t,n}),
\end{dmath}

where, $\lambda$ shows how sensitive the objective function is to changes in the value of $\mathcal{A}_{t,n} \geq A_{\text{threshold}}$, the accuracy constraints. By using this approach, we are able to ensure that minimizing the objective function will only occur if the accuracy requirements are achieved. Otherwise, OP is penalized and stops optimizing more during $\mathcal{A}_{t,n} < A_{\text{threshold}}$. The fixed $\lambda$ creates a lack of convergence and over- and under-penalization constraint violation during latency optimization. To address these issues, we use dual gradient descent to dynamically update the $\lambda$ as mentioned in (\ref{dual_gradient}):

\begin{equation}\label{dual_gradient}
\lambda \leftarrow \lambda + \eta (A_{\text{threshold}} - \mathcal{A}_{t,n}),
\end{equation}
where $\eta$ is the learning rate, which is initially set to 0.01 with a weight $\lambda$ of 1. The $\eta$ dynamically adjusts $\lambda$ during constraint violation. So $r_t^{\text{A2C}}\left \{{\mathbb{S}^\psi_i \mid \mathbb{S}^*_i, \mathbb{B}_{t,n}}\right \}$ of the training of A2C is determined by offloaded group of pixel sets to the server and is given in (\ref{Reward}) by 

\begin{equation}\label{Reward}
r_t^{\text{A2C}}\left \{{\mathbb{S}^\psi_i \mid \mathbb{S}^*_i, \mathbb{B}_{t,n}}\right \} 
= \min _{\forall i_{\mathbb{S}^*_i} \in [1,Z]} 
T_{t,n}.
\end{equation}

We incorporate the reward into the temporal difference learning $(\delta_t)$ error for the critic and actor loss as mentioned in \eqref{criticloss}:

\begin{equation}\label{criticloss}
\delta_t = r_t^{\text{A2C}}\left \{{\mathbb{S}^\psi_i \mid \mathbb{S}^*_i, \mathbb{B}_{t,n}}\right \} 
+ \gamma V^*({\Phi}_{t+1}) - V^*({\Phi}_t),
\end{equation}
where, $V^*({\Phi}_{t+1})$ is the update estimated state of ${\Phi}_t$ that is the old estimated state of $V^*({\Phi}_t)$. The critic loss is evaluated by squaring the $(\delta_t)$ to ensure how well the $V^*({\Phi})$.

While the actor loss is based on the advantage function to regulate the policy $(\pi_{\theta}(a_t \mid {\Phi}_t))$ and map the states $({\Phi}_t)$ to the actions $(a_t)$ as given as 
$-\log \pi_{\theta}(a_t \mid {\Phi}_t) \cdot \delta_t$. $-\log \pi_{\theta}(a_t \mid {\Phi}_t)$ takes action under the current policy $\pi_{\theta}$ for the state ${\Phi}_t$ and takes the product of the likelihood of action with $\delta_t$ to update the policy using \eqref{LossA2C}:

\begin{equation}\label{LossA2C}
    \mathcal{L}_{loss} = -\log \pi_{\theta}(a_t \mid {\Phi}_t) \cdot \delta_t 
    + \mathcal{W}_c \cdot \left( \delta_t \right)^2,
\end{equation}
where $\mathcal{W}_c $ shows the weighted factor for the critic loss. $\mathcal{L}_{loss}$ defines that the actor is maximizing the advantage and critic is minimizing the $\delta_t$ error with current updates within the constraints defined by A2C reward $r_t^{\text{A2C}}\left \{{\mathbb{S}^\psi_i \mid \mathbb{S}^*_i, \mathbb{B}_{t,n}}\right \}$.

\subsubsection{Policy}
The policy $\pi_\theta(a_t|{\Phi}_t)$ represents the A2C actor network that maps the CMDP state ${\Phi}_t$ to a probability distribution over actions $a_t$. Each action corresponds to selecting a bounding box $\mathbb{B}_{t,n}$, which 
defines the ROI $f_{t,n}$ to be transmitted. The policy parameters $\theta$ are updated using the actor loss in (\ref{LossA2C}), which leverages the advantage signal $\delta_t$ to encourage actions that reduce latency $T_{t,n}$ while maintaining 
the required accuracy $\mathcal{A}_{t,n}$.

\section{Experimental Work}
\subsection{Implementation of GCN-assisted A2C Model}
This section presents an overview of the results for onboard detection of intruding UAVs through the proposed GCN-assisted A2C model. The proposed GCN-assisted A2C model is built on a 64-bit Ubuntu 22.04 Linux workstation using Python 3.9 and PyTorch with a GPU of A300. In this study, we used YouTube videos for UAV-to-UAV monitoring. There are a total of 100 clips, all of which are 240 to 1080 resolution videos. These videos feature a variety of backgrounds, such as clouds, buildings, mountains, and sea, as well as fast movement, out-of-focus, and a range of sizes. The UAVs seen in each video come in a range of sizes, from large to small. We selected UAVs with a size range of 300mm to 1200mm to train and validate our proposed model. Furthermore, the UAVs cruise at speeds between 50 and 100 miles per hour, occasionally stopping altogether. We used all the validation videos from the YouTube stream for model training, testing and validation. Specifically, we train the model on 70 of the videos, using 15 of the remaining for testing and the remaining 15 for validation.

The GCN-assisted A2C is trained for 1 iteration per epoch to find the correlation between the different regions in every frame, and the GCN-assisted A2C is trained for 100 epochs to extract the pixel feature-correlated regions to send it to the server. GCN-assisted A2C is analyzed every 5 epochs using a validation function during the training process. This function finds the training loss and the number of iterations at which convergence occurs. The GCN-assisted A2C model with the lowest cost, denoted as the sub-optimum, is then used to assess the test data.

The GCN-assisted A2C used multiple layers that are organized into the GCN layers and multi-layer perceptron layers. The GCN layers module consists of three GCN multi-head layers, each containing two GCN layer edge softmax heads. The first GCN multi-head layer utilizes heads that apply a linear transformation $f$ to convert input pixel features into output-correlated pixel features. Additionally, another linear transformation is used to convert input pixel features into output pixel feature-correlated features. The second layer undergoes a transformation where 128 input features are converted to 64 output pixel features, and 128 input pixel features are converted to 1 output pixel feature in a region that is correlated. The third layer of the GCN model performs a similar transformation by changing 256 input layer pixel features into 64 output pixel features and 256 input pixel features into 1 output pixel feature-correlated group of regions. The A2C module has two linear layers, in which the first layer maps 256 input pixel features to 128 output pixel features. For the final prediction, the second layer maps 128 input features to output feature-correlated group regions. It does this by using (\ref{x_c}) and (\ref{B_i}) for the centroid.

The calculation of the policy gradients after multiplying by the benefits is known as policy loss. The Unmanned Aerial Vehicle (UAV) activities undergo training in the Advantage Actor-Critic (A2C) module for a total of 1500 steps, using a discount factor of 0.99. The loss function is calculated by taking the average of the squared differences between the estimated values and the actual returns. The loss calculation assigns a coefficient of 0.5 to the value function.

\subsection{Ablation Study}
\begin{figure}[htbp!]
    \centering
    \hspace*{-1em}
    \includegraphics[scale=0.58]{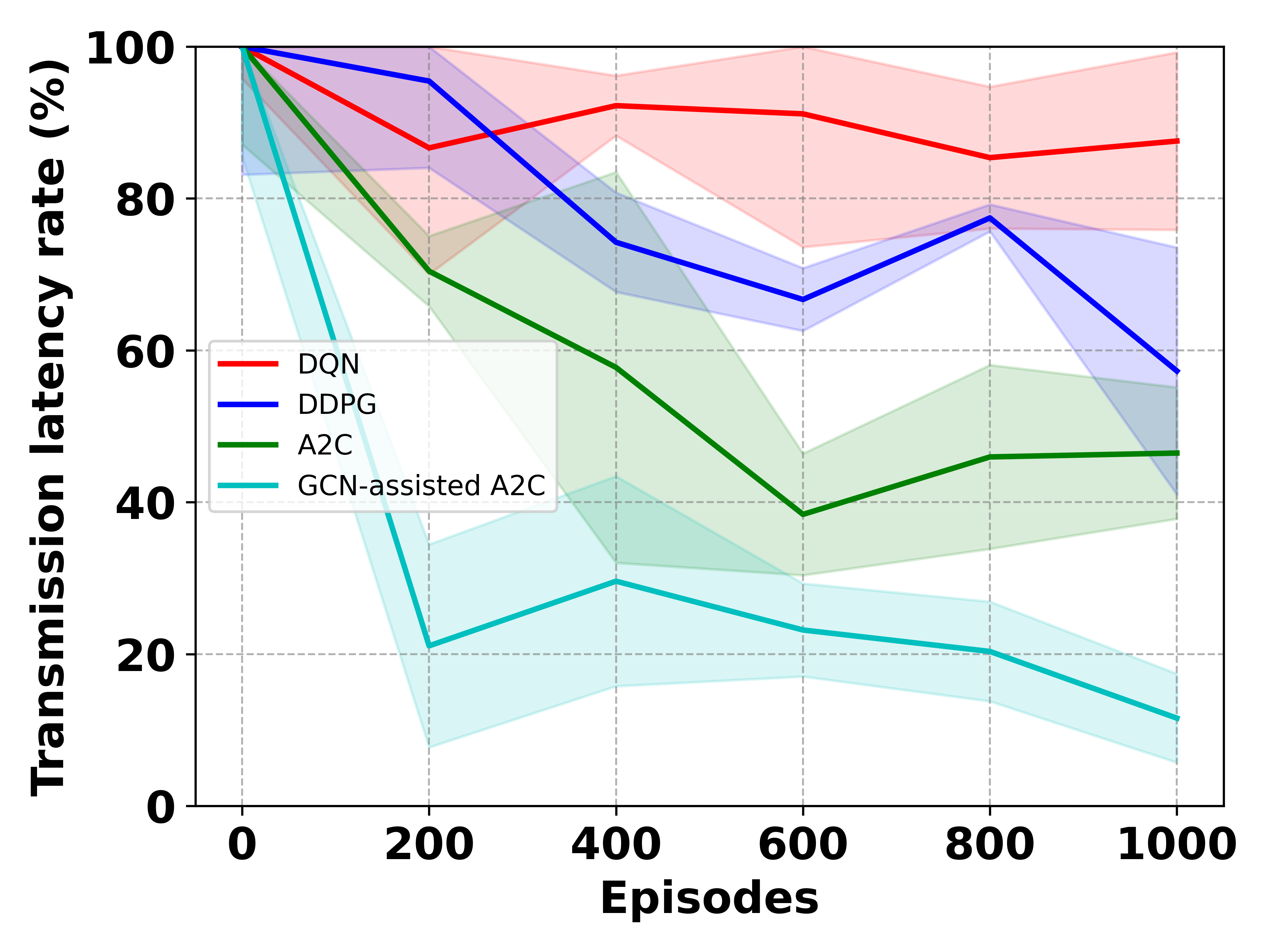}
    \caption{The ablation study of different RL algorithms performance using loss using data that have across different sizes of UAVs and dynamic movements with different environmental effects.}
    \label{Ablation_Study}
\end{figure}

In Fig. \ref{Ablation_Study}, we can see how well the proposed GCN-assisted A2C system model and different DRL algorithms work in terms of transmission latency rate during training. We found that the A2C, DDPG, and DQN models mostly do the same thing, which is to minimize transmission latency. However, the synchronous nature of the A2C model leads to higher performance than the deterministic nature of the DDPG and DQN models. The frame area's importance and the dynamical changes in the UAV's size due to different movements make A2C more stable than the DDPG and DQN models. In addition, cropping the frame area makes a clear action space, which might be one reason why A2C is generally more useful than DDPG. The frame area cropping creates priority fluctuation that leads A2C to better decision-making. It's hard for the DDPG and DQN models to make the best decision when there are a lot of different environmental factors and frame area cropping variability problems. However, to adopt and find feature correlations and dependencies between different regions of the frame still reduces the performance of A2C. We proposed a GCN-assisted A2C system model, which ensures and preserves critical contextual features of groups of neighboring pixels in regions where A2C lacks the ability to deal with such dependencies. A2C processes and transmits a single sub-area frame entity and makes it difficult to find the nuanced interplay between different sub-areas of frame parts or sizes of UAVs, variations, or environmental factors. Moreover, GCN assists the A2C to extract informative features of the UAV using aggregation and propagation of features across graph nodes to make better predictions. Similarly, the GCN-assisted A2C system model reduces the high dimensionality focus to the specific interdependent area to efficiently handle energy constraints or bandwidth limitations. Without GCN, A2C struggles with scalability, and it leads to high computation, and might be complex architectures needed. In addition, GCN-assisted A2C is more robust to noises or missing pixel feature information to aggregate neighboring nodes feature information to improve predictions and reduce the transmission latency. If a portion of a frame is corrupt or unclear with environmental effects, then GCN leverages the hidden feature pixel information of the surrounding group of pixels, improves the transmission latency, and preserves the prediction performance constraints.             

\begin{figure}[htbp!]
\centering
\hspace*{0em}
{\includegraphics[scale=1.1]{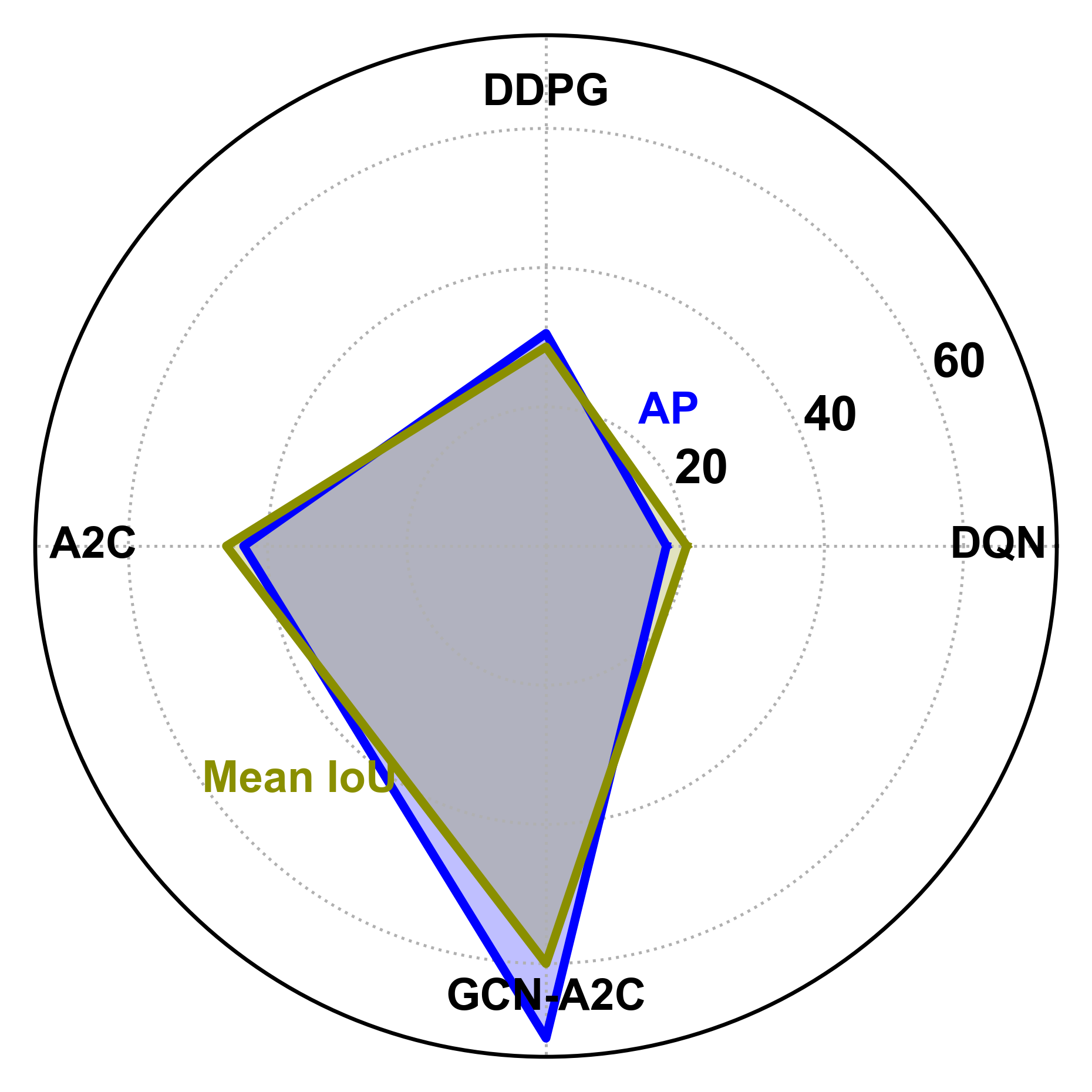}}
\caption{With the proposed GCN-assisted A2C and different DRL algorithms, we can check the average (30FPS) IoU and tracking AP accuracy.}
\label{Latency_AP_meanIoU} 
\end{figure}

Moreover, in the ablation study, we used different DRL models (DQN, DDPG, and A2C) and studied required evaluation metric performance, like latency for each video frame transmission, overall AP, and mean IoU. The latency of the GCN-assisted A2C is much lower than that of the other benchmark DRL algorithms. The overall FPS transmission latency of the GCN-assisted A2C is 45 ms, which is much lower than the A2C without GCN, which has an FPS transmission latency of 170 ms. The GCN-assisted A2C achieved optimal latency due to the focus on the feature-correlated group of pixel regions to recommend to the A2C for optimal action. Moreover, Fig. \ref{Latency_AP_meanIoU} shows the AP and mean IoU of the GCN-assisted A2C and the three variants of the ablation study. The GCN-assisted A2C is the most effective across all the algorithms to transmit the optimal area of the frame that is utilized by the edge server. The GCN-assisted A2C achieved (AP @ 70.72\%) and (mean IoU @ 60.30\%) to improve (AP @ 62.80\%) and (mean IoU @ 30.60\%) over A2C. While A2C achieved (AP @ 43.44\%) and (mean IoU @ 45.91\%), which is higher in ratio than (AP @ 42.20\%) and (mean IoU @ 60.50\%) over the DDPG algorithm. Similarly, DDPG achieved (AP @ 30.55\%) and (mean IoU @ 28.61\%), and DQN with a very low score of (AP @ 17.30\%) and (mean IoU @ 20.21\%).


\subsection{State-of-the-art (SOTA) comparisons}
\begin{figure}[htbp!]
    \centering
    \hspace*{-1em}
    \includegraphics[scale=0.58]{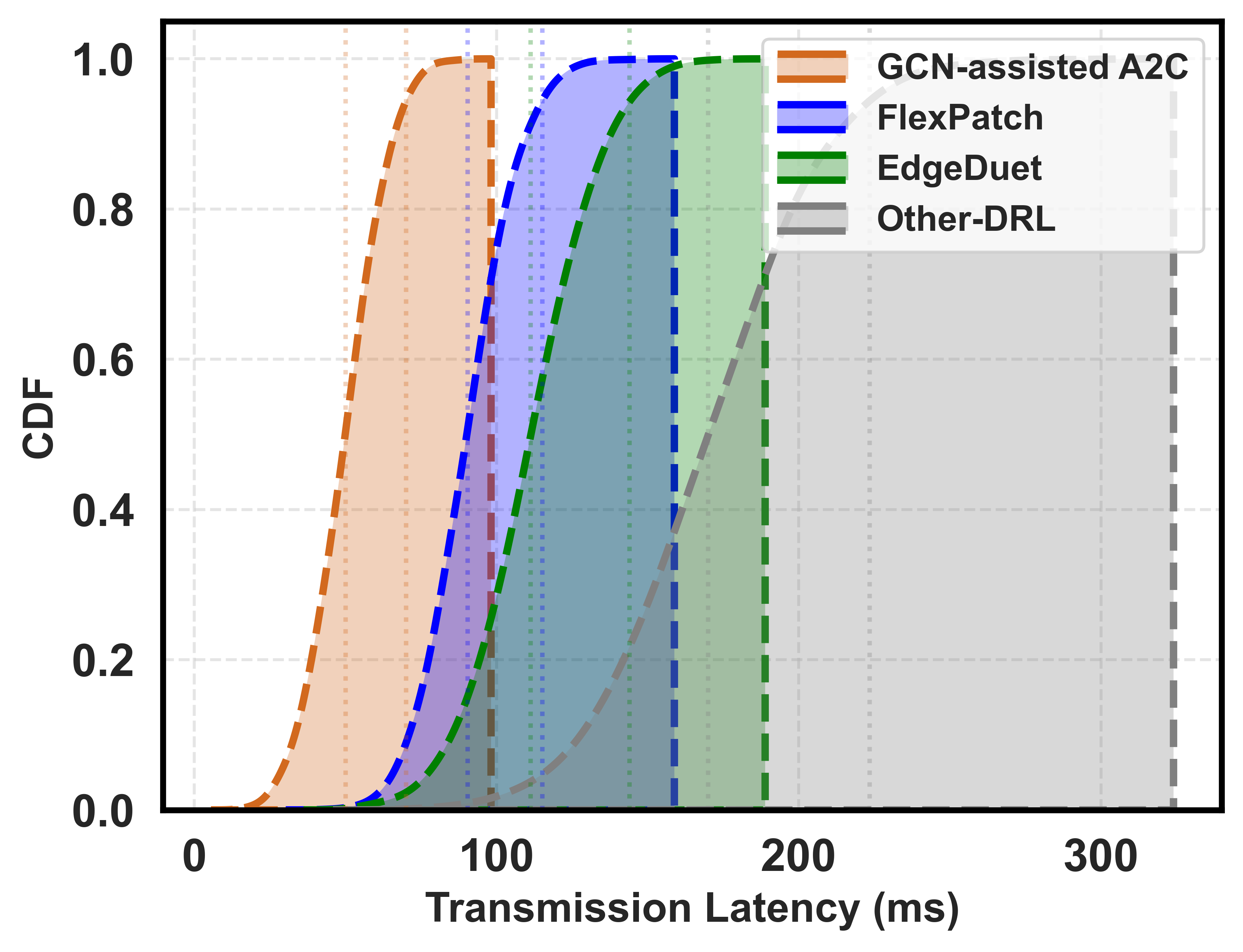}
    \caption{The effectiveness of GCN-assisted A2C over SOTA models to optimize the transmission latency.}
    \label{SOTAmodels}
\end{figure}

Fig. \ref{SOTAmodels} shows the statistical performance of different state-of-the-art (SOTA) models, i.e., FlexPatch, EdgeDuet, and other DRL, to visualize the transmission latency in the range of cumulative distributed function (CDF) parameters. We found that the proposed GCN-assisted A2C had the best transmission latency of 50 ms with a standard deviation of 12 ms when compared to other SOTA models. FlexPatch came in second with a transmission latency of 90 ms and a standard deviation of 15 ms. Meanwhile, EdgeDuet achieved a video frame transmission latency of 110 ms, and its standard deviation is 20ms. In contrast, other DRL models achieve a video frame transmission latency of 177 ms, with a standard deviation of 33 ms. We can see from the CDF of the GCN-assisted A2C that the model works better for achieving the best video frame transmission latency across a range of communication factors.

\begin{figure*}[htbp!]
\centering
\hspace*{0em}
\subfloat[]{\includegraphics[width=1.8in]{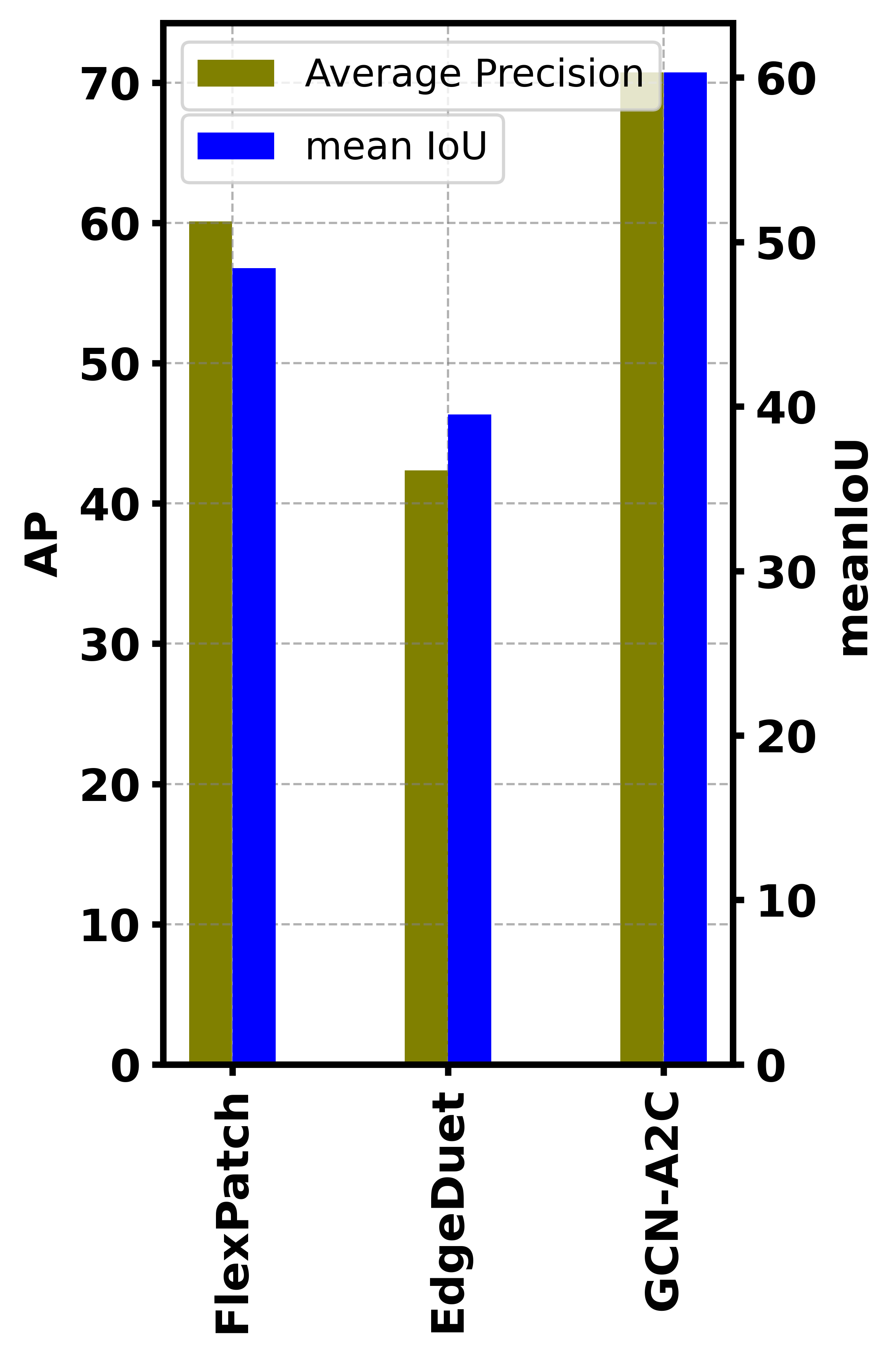}} 
\subfloat[]{\includegraphics[width=1.8in]{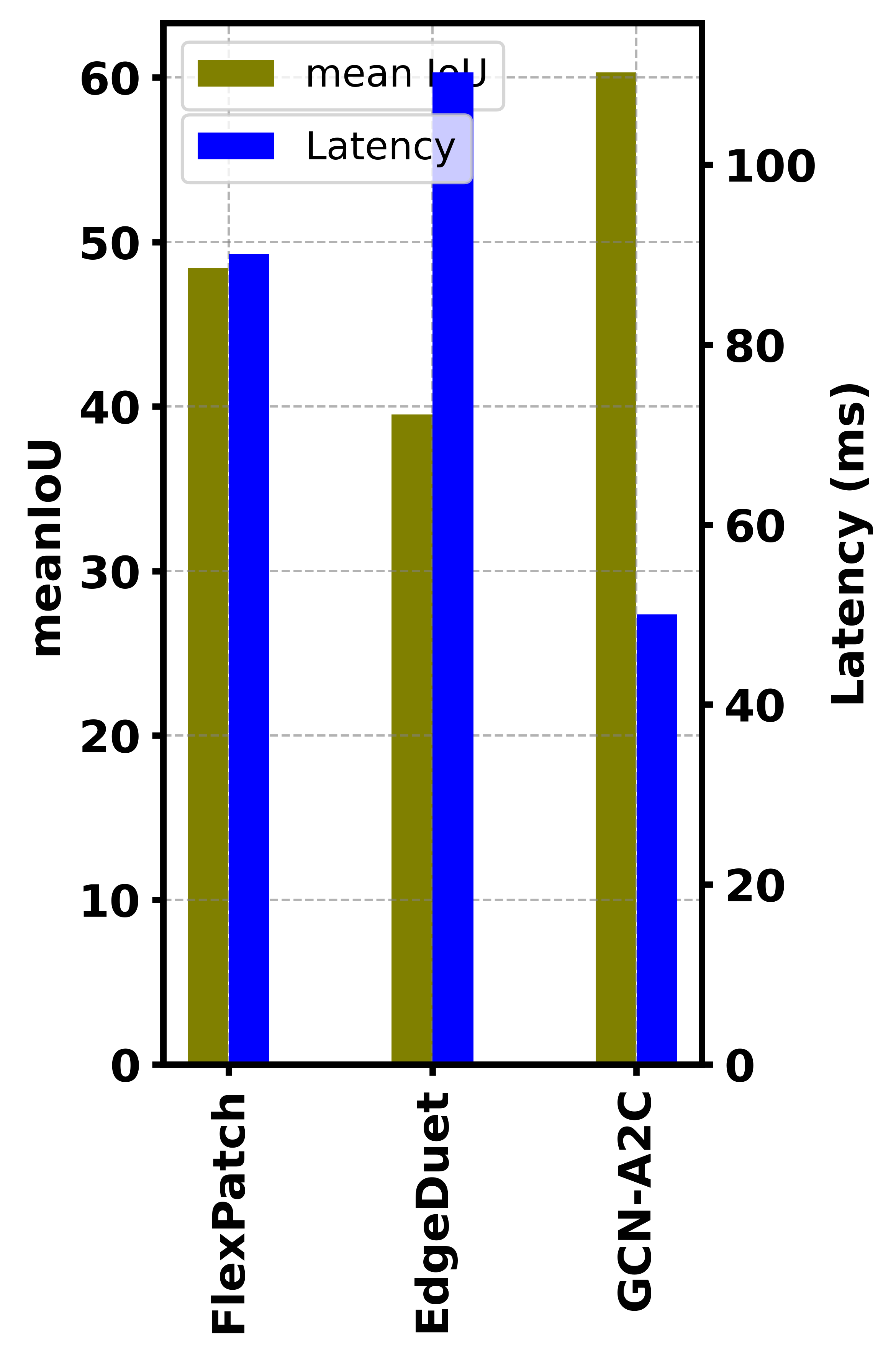}}
\subfloat[]{\includegraphics[width=1.8in]{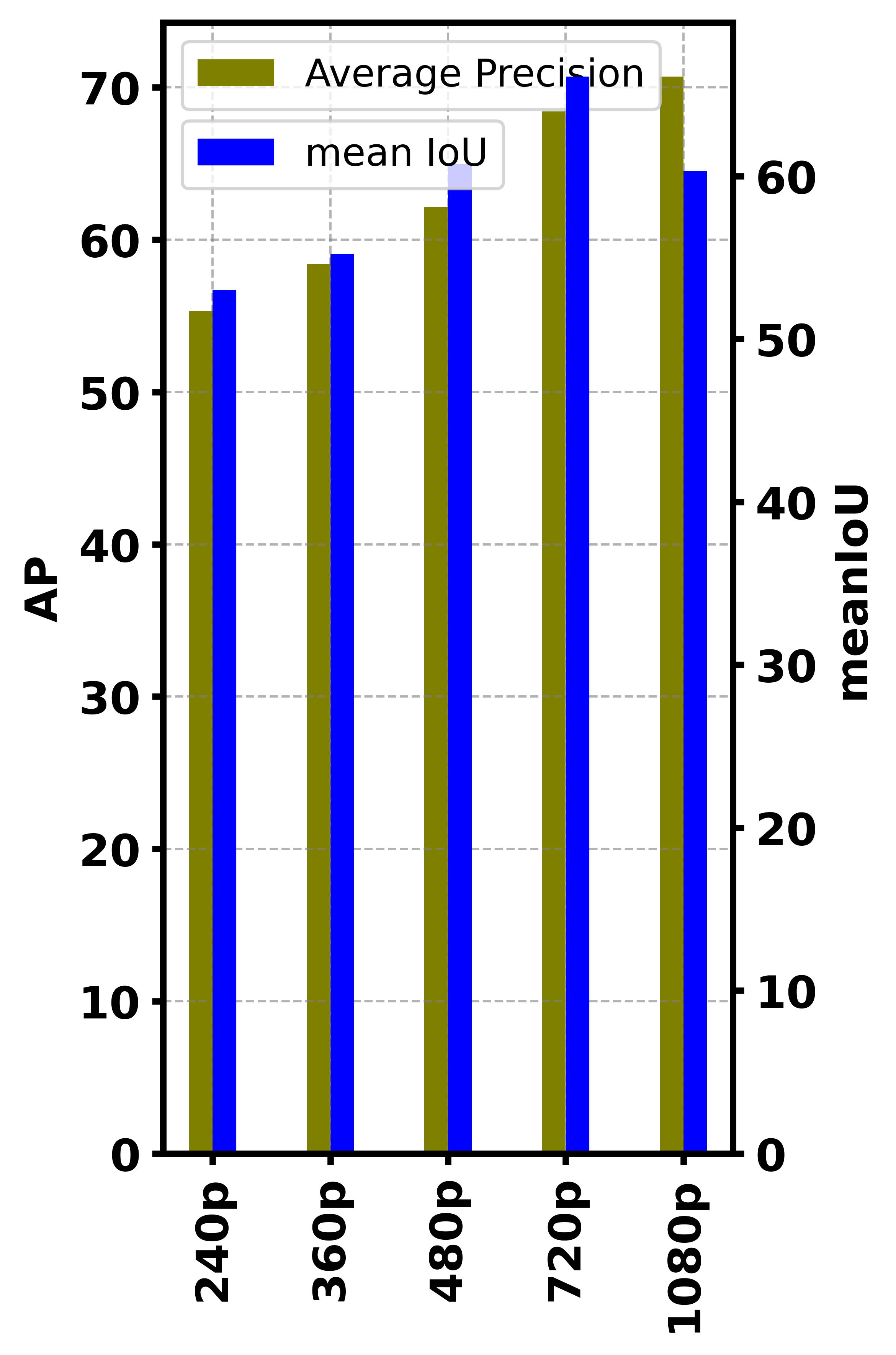}}
\caption{Performance of different models under UAV-to-server video sub-area of frame transmission. In (a), the AP and meanIoU accuracy of finding suspicious UAVs from end-to-end are shown. In (b), the meanIoU over end-to-end latency is compared. And in (c), the performance of GCN-assisted A2C is shown over different input video frame resolutions.}
\label{SOTA_AP_IoU_Latency_Resol} 
\end{figure*}

At the same time, we examined how well the GCN-assisted A2C worked regarding transmission delay, AP, and meanIoU accuracy when compared to FlexPatch and EdgeDuet, using different video resolutions as shown in Fig. \ref{SOTA_AP_IoU_Latency_Resol}. In Fig. \ref{SOTA_AP_IoU_Latency_Resol}(a), we show how accurately we can find suspicious UAVs from start to finish, using AP and meanIoU accuracy. Similarly, in Fig. \ref{SOTA_AP_IoU_Latency_Resol}(b), the mean IoU over end-to-end latency is compared. The GCN-assisted A2C model outperforms previous models. For example, FlexPatch AP accuracy performance is 60.10\% with meanIoU accuracy results of 48.42\% and achieved minimum latency results of 90.10 ms. The FlexPatch adds overhead when scenes change rapidly or unpredictably, and it is highly challenging to select the optimal context-dependent features. EdgeDuet achieved AP accuracy of 42.35\% with meanIoU accuracy results of 39.52\% and achieved minimum latency results of 110.25ms. This model also uses many high-resolution tiles at the edge server and is unsuitable for real-time or low-bandwidth scenarios due to offloading frame tiles to the edge server. The GCN-assisted A2C model is better at handling quickly changing or unpredictable scenes because it uses groups of pixel areas as nodes and their connections to understand how they depend on each other. The GCN-assisted A2C model reached the best accuracy of 70.72\% AP, with meanIoU accuracy results of 60.30\%, and improved the transmission delay to 50ms. The GCN-assisted A2C model works well in real-time or low-bandwidth situations because it can handle parts of the frame by grouping together related pixel areas. 

Fig. \ref{SOTA_AP_IoU_Latency_Resol}(c) shows the AP accuracy and meanIoU of the GCN-assisted A2C when transmitting videos of different frame sizes. The GCN-assisted A2C consistently achieves higher AP meanIoU accuracy at the server side. With the decrease in frame resolutions, GCN-assisted A2C accuracy drops upto 50 AP, which is our desired threshold to have a minimum of 50 AP at the server side. 

\subsection{Evaluation Results}
\begin{figure*}[htbp!]
    \centering
    \hspace*{-1em}
    \includegraphics[scale=0.08]{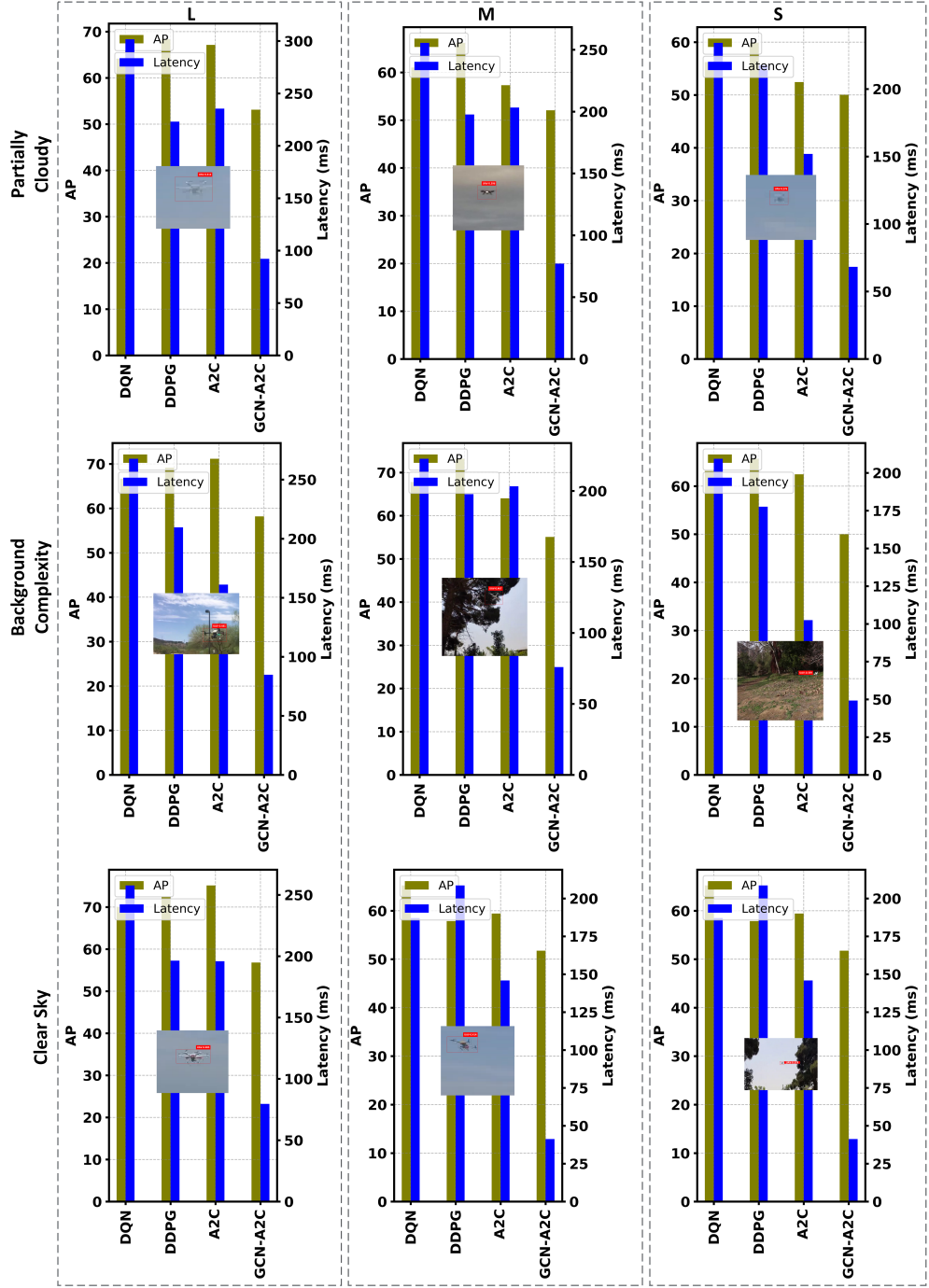}
    \caption{We evaluated three different sizes of UAVs—small (S), medium (M), and large (L)—under various conditions, such as clear sky, complex background, and partially cloudy.}
    \label{Eval_Results}
\end{figure*}

We have qualitatively and quantitatively shown the expected trade-off between AP and latency improvement for three different UAV sizes under various circumstances. Each environment has three visual situations, with each subplot representing the AP and transmission latency. The blue bars show latency, whereas the olive-colored bars represent AP. Together, they exhibit both precision and efficiency. Overlaid visuals in each subfigure illustrate corresponding visual inputs related to the environmental condition. This shows how complex the agents' experiences are. Fig. \ref{Eval_Results} shows that smaller UAVs, similar backgrounds, and partial cloudiness can affect latency and reduce server-side AP. These unique features have a major effect on the efficiency of transmission in the other DRL models. The proposed GCN-assisted A2C keeps an AP above 50 while ensuring the best transmission speed in all conditions and sizes of UAVs during testing of the algorithms in three different conditions—cloudy, complex, and clear sky. This hypothetical situation motivates the advancement of smarter feature-correlated region prediction algorithms. The results indicate that GCN-assisted A2C consistently does better than the basic models in every situation, achieving a much higher AP with less delay. This benefit is especially evident in cloudy or messy situations, such as cloudy and complex situations, where classic models like DQN and DDPG do not perform well. GCN-assisted A2C, on the other hand, is robust to environmental changes and visual effects, making it more stable and suitable for a wider range of applications.  In addition, GCN-assisted A2C continues to outperform its competitors in ideal situations such as clear skies. This demonstrates that it performs well in both challenging and normal circumstances. These findings show that GCN-assisted A2C could be effective for making real-time predictions based on vision when transmission speed and precision are important.

\section{Conclusion}
This article proposed a GCN-assisted A2C model that uses data from the UAV detection nano YOLO algorithm, focusing on the center of the detected object and the surrounding pixels. The design caters to scenarios in which on-board UAV detection accuracy is less than 50. We designed the detector model to identify suspicious UAVs in no-fly zones and pinpoint their location, utilizing the onboard UAV cameras. Next, we proposed using a GCN-assisted A2C to enhance the significance of certain pixel areas in the images by examining how each area connects to the center of the UAV, creating groups based on these connections, and calculating a total score with A2C. The nano detector model results emphasize that the GCN-assisted A2C model requires transmitting the frames to the server or not. We show that using the GCN-assisted A2C model can minimize the transmission latency and optimize until the AP is more than 50 on the server side. We conducted studies using authentic on-board UAV camera data and with different DRL models for the ablation study and SOTA models. Our GCN-assisted A2C model shows improved transmission latency and AP by using GCN to group related pixels and understand hidden pixel relationships, which helps the A2C make better decisions in different situations like clear skies, complex backgrounds, and partial clouds.

\section*{Acknowledgments}


\bibliography{Bibfile.bib}

\vspace{11pt}

\vspace{11pt}

\vfill

\end{document}